\pdfoutput=1

\documentclass[11pt]{article}

\usepackage[final]{acl}

\usepackage{times}
\usepackage{latexsym}

\usepackage[T1]{fontenc}

\usepackage[utf8]{inputenc}

\usepackage{microtype}

\usepackage{inconsolata}

\usepackage{graphicx}
\usepackage{pifont}
\usepackage{booktabs,amsfonts,dcolumn}
\usepackage{amsmath,amsthm,amssymb,bm,stmaryrd,bbm}
\usepackage{footmisc}\interfootnotelinepenalty=10000

\usepackage{float}
\usepackage{array}
\usepackage{mathtools}
\usepackage[export]{adjustbox}
\usepackage{xcolor}
\usepackage{natbib}
\usepackage{enumitem}
\usepackage{amsmath}
\usepackage{multirow}
\usepackage{enumitem}
\usepackage{url}
\usepackage[capitalise]{cleveref}
\usepackage{tabularx}
\newcolumntype{C}{>{\centering\arraybackslash}X}
\newcolumntype{d}[1]{D..{#1}}

\usepackage{subcaption}
\usepackage{algorithm}
\usepackage{algpseudocode}

\graphicspath{ {./images/} }

\title{CoTAR: Chain-of-Thought Attribution Reasoning with Multi-level Granularity}

\author{Moshe Berchansky \qquad Daniel Fleischer \qquad Moshe Wasserblat \qquad Peter Izsak \\
Intel Labs \\
 \tt \{moshe.berchansky,daniel.fleischer,moshe.wasserblat,peter.izsak\}@intel.com \\
}

\begin{document}

\maketitle
\setlength{\abovedisplayskip}{2pt}
\setlength{\belowdisplayskip}{2pt}

\begin{abstract}

State-of-the-art performance in QA tasks is currently achieved by systems employing Large Language Models (LLMs), however these models tend to hallucinate information in their responses.
One approach focuses on enhancing the generation process by incorporating attribution from the given input to the output. However, the challenge of identifying appropriate attributions and verifying their accuracy against a source is a complex task that requires significant improvements in assessing such systems.
We introduce an attribution-oriented Chain-of-Thought reasoning method to enhance the accuracy of attributions. This approach focuses the reasoning process on generating an attribution-centric output.
Evaluations on two context-enhanced question-answering datasets using GPT-4 demonstrate improved accuracy and correctness of attributions. In addition, the combination of our method with finetuning enhances the response and attribution accuracy of two smaller LLMs, showing their potential to outperform GPT-4 in some cases.\footnote{Our code is publicly available for reproduction: \url{https://github.com/mosheber/cotar.git}}

\end{abstract}

\section{Introduction}

Text generation from sources, such as information obtained through retrieval, is a key aspect of grounded question answering. One way to ensure the credibility of these models is through attributed text generation. This method pairs the generated text with supporting evidence, enhancing the model's trustworthiness and allowing for easier detection of errors. 


The pairing process can be done in several levels. Firstly, we can pair each sentence in our response with a set of passage identifiers \cite{gao2023enabling}. On a model detailed level, individual spans in the answer can correspond to a specific passage, hence attributing specific portions to various passages \cite{Schuster2023SEMQASM}. These spans are designed to be direct copies from the passages, assuring that the content provided is fully supported. Since not all the spans are copied, the model has a sufficient degree of flexibility to produce both a coherent and factually correct answer.

\begin{figure}[t]
\centering
\includegraphics[width=0.65\columnwidth]{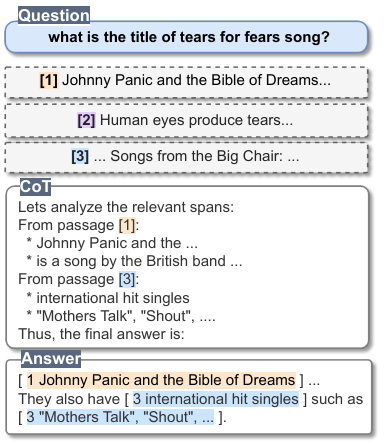}
    \caption{Usage of CoT for attribution-based answers. We either instruct the model, using fewshot examples, or finetune the model, to produce a detailed list of the salient information from each passage. Each entry can be either on the passage, sentence, and even span level. Finally, the model produces a coherent and faithful answer.}
\end{figure}

However, these approaches fail at times in two primary ways to cite accurately. On one hand, the model might focus only a very specific portion of the input, therefore missing relevant sections when it constructs the answer. On the other hand, it might cite too many passages, some of which might not be relevant to the answer.

To solve these, we propose utilizing a Chain-of-Thought approach (CoT)~\cite{Wei2022ChainOT}, which we denote as CoTAR, which allows the model to perform reasoning over the input passages before generating the output. In our approach, we instruct the model to extract relevant information from the passages as well as specify the form of attribution to generate. The information can be in the span, sentence, or passage level. We inspect how the various CoT methods affect the model's output, in each of the citation levels. In addition to few-shot instruction, we also finetune various models using our method, resulting in models that are competitive with and even outperform GPT-4 in some cases. 

We can summarize our contributions in this paper as follows:

\begin{itemize}[leftmargin=*,noitemsep]\vspace{-5pt}
    \item We perform rigorous measurements of both the answer quality and citation quality, across multiple models and citation levels.
    \item We show that utilizing CoT reasoning improves the ability of an LLM to produce both better quality answers, and more precise and faithful citations from the source, demonstrated on multiple models.
    \item We demonstrate that using by finetuning, smaller models can be competitive with or outperform GPT-4 in some cases in answer and citation quality metrics.
\end{itemize}


\section{Method}
\label{sec:method}

\subsection{Attribution-Oriented Question Answering}
\label{sec:cite_task_explain}

Attribution-oriented question answering can be defined as a task where, given a query and a set of relevant contextual resources, the objective is to accurately answer the question while attributing specific portions or the entire answer to the appropriate contextual sources.

We can identify three levels of attribution with different granularity levels: \textbf{Span}, \textbf{Sentence} and \textbf{Passage}. Attribution levels refer to how citations are displayed within an answer, indicating whether complete passages, individual sentences, or specific text segments are credited back to their original sources. 
Detailed examples for each level are available in \cref{tab:cite_level_example} in the appendix. 




\subsection{Chain-of-Thought Attribution Reasoning}
\label{sec:cot_method_explain}


We propose a multi-step CoT~\cite{Wei2022ChainOT} reasoning scheme with varying levels of attribution, similar to the three levels presented in \cref{sec:method}, hypothesizing that this could encourage the model to generate more accurate answers. The process involves identifying the most crucial aspects of the given context for answering the question, by incorporating direct citations to the referenced parts. We denote the process of granular attribution as CoT \textit{attribution guidance}, or CoT method for short. 
The three levels are as follows:
\begin{itemize}[leftmargin=*,noitemsep]\vspace{-5pt}
    \item \textbf{Span Guidance}: Produce the relevant spans of information per passage.
    \item \textbf{Sentence Guidance}: For every passage, write sentences that summarize how the passage answers the question.   
    \item \textbf{Passage Guidance}: State which passages are relevant for the question.
\end{itemize}

We explore all combinations of CoT methods with citation levels, and asses how each one effects the generated answers. A example of all three levels of attribution-guidance is in \cref{tab:cot_level_prompts} in the appendix. 



\begin{table*}[!ht]
\small
\centering
\begin{tabular}{ccrrrrrrrrr}
\toprule
\multirow[c]{2}{*}{Level} & \multirow[c]{2}{*}{CoT} &  \multicolumn{4}{c}{Answer Quality} & \multicolumn{5}{c}{Citation Quality} \\
 \cmidrule(lr){3-6} \cmidrule(lr){7-11}
 &  & BERT & HEM & RL & \textit{Avg.} & ALCE F1 & CSCA & DOC F1 & SEM-F1 & \textit{Avg.} \\
\midrule
\multirow[l]{4}{*}{Span} & None & 91.71 & \bfseries 67.58 & 66.08 & 75.12 & 77.06 & 81.15 & 85.33 & 67.57 & 77.78 \\
 & Span & \bfseries 92.06 & 66.95 & 68.55 & \bfseries 75.85 & 78.83 & \bfseries 89.88 & \bfseries 86.54 & \bfseries 72.23 & \bfseries 81.87 \\
 & Sent. & 91.87 & 66.88 & \bfseries 68.76 & 75.84 & \bfseries 84.76 & 69.69 & 86.38 & 70.52 & 77.84 \\
 & Pass. & 91.91 & 66.89 & 67.67 & 75.49 & 79.21 & 86.94 & 86.49 & 69.87 & 80.63 \\
\midrule
\multirow[l]{4}{*}{Sent.} & None & 91.47 & 62.48 & 62.18 & 72.04 & 82.79 &-& 85.32 & 59.20 & 75.77 \\
 & Span & 91.92 & 62.06 & 64.76 & 72.91 & 82.39 &-& \bfseries 86.71 & 61.74 & 76.94 \\
 & Sent. & \bfseries 91.96 & \bfseries 64.59 & \bfseries 67.36 & \bfseries 74.64 & \bfseries 86.42 &-& 86.51 & \bfseries 64.44 & \bfseries 79.12 \\
 & Pass. & 91.69 & 61.19 & 64.42 & 72.43 & 84.25 &-& 85.92 & 61.79 & 77.32 \\
\midrule
\multirow[l]{4}{*}{Pass.} & None & 91.55 & 61.13 & 62.76 & 71.81 & \bfseries 75.38 &-& 85.19 & \bfseries 44.31 & \bfseries 68.29 \\
 & Span & \bfseries 91.90 & 61.66 & 64.70 & 72.75 & 71.60 &-& 86.17 & 43.41 & 67.06 \\
 & Sent. & 91.84 & \bfseries 62.43 & \bfseries 65.32 & \bfseries 73.20 & 70.63 &-& \bfseries 86.41 & 42.59 & 66.54 \\
 & Pass. & 91.69 & 60.40 & 63.88 & 71.99 & 73.28 &-& 85.64 & 44.12 & 67.68 \\
\bottomrule
\end{tabular}

\caption{Results of GPT-4 on the QuoteSUM dataset. The results are categorized by response attribution (level), and for each level, every CoT approach is demonstrated. The best results per level are marked in \textbf{bold}.}
\label{tab:gpt4_res_few}
\end{table*}

\section{Evaluation Metrics}

In order to evaluate the answers, we specify metrics of two types. \textbf{Answer Quality}, measuring similarity between the predicted answer and the gold answer. \textbf{Citation Quality}, assessing the similarity between the cited text, the cited passage, and the citations present in the gold answer.


For answer quality , we use the n-grams based method ROUGE-L (RL), and the semantic method BERTScore (BERT)~\cite{bert-score}. For hallucination evaluation, we use the HEM\footnote{\url{vectara/hallucination_evaluation_model}} model, which was finetuned on NLI datasets, and predict whether two texts are factually consistent. For citation quality, we measure the quality of the cited content; we propose specific metrics for each of the citation levels. We note that span citations must match the source passages precisely, while sentence and passage attributions are not required to do so. We use the \textit{SEM-F1} metric proposed by \citet{Schuster2023SEMQASM} for n-gram token level similarity between the cited content. In addition, we include the Citation Precision/Recall introduced in \citet{gao2023enabling}, and combine them into \textit{ALCE F1}, by using the harmonic mean. We also measure the F1 score over passage indices between the cited passages and the passages cited by the expected answer, denoted as \textit{DOC F1}, and the \textit{Correct Span Citation Attribution (CSCA)}  which indicates whether a predicted span is a direct span from the attributed passage.

\section{Experimental Setup}



In order to properly manage our experimental setup, we utilized RAGFoundry, as introduced in \citet{fleischer2024ragfoundryframeworkenhancing}. This allowed us to create the data preprocessing flows, train our models, conduct inference, and evaluate on the various metrics in a customizable and seamless manner. Our codebase provides a detailed description of how to reproduce the results using the custom steps we have incorporated with the framework.

\subsection{Datasets}
\label{sec:dataset_names}


We use the QuoteSum (QSUM) \cite{Schuster2023SEMQASM} dataset, which contains semi-extractive answers written by humans, for natural questions and generated ones. In addition, we include MS MARCO (MS)\footnote{\url{https://huggingface.co/datasets/ms_marco}} \cite{Campos2016MSMA}, which is a crowd sourced dataset of responses to queries from Bing. Each example include 10 passages which can be relevant or irrelevant to answering the question.

\subsection{Span Dataset Preparation}

In order to include span citations in datasets such as MSMARCO, we can create span-based citation labels out of passage-based labels. We do so by finding the longest common substrings between the input passages and the answer, and choosing those that contain named entities, thus ensuring the semantic relevance of the cited spans. The answer is then formatted as proposed by \citet{Schuster2023SEMQASM}. For more details about the process, we refer the reader to \cref{alg:two}.


\subsection{Models}

\label{sec:model_details_setup}

We use GPT-4 \cite{openai2024gpt4} for evaluating our CoT method. We use the Mistral 7B \cite{jiang2023mistral} decoder model and the Flan-T5 XXL \cite{chung2022scaling} encoder-decoder as representative small-sized models in our further finetuning evaluations.




\begin{algorithm}[H]
\caption{Generating span attirubtions from a passage-labeled dataset.}\label{alg:two}
\begin{algorithmic}[1]
\State Input: Answer $a$, passages $\{p_i\}$
\State Extract named entities $\text{NER}(a)$ using spaCy \cite{spacy2}.
\State Find common \textit{substrings} $CS(a, p_i)$ between every passage $p_i$ and answer $a$.
\State Each substring $sc \in CS(a, p_i)$: contains $sc_t$ (string text) and passage index $sc_p$.
\State Sort substrings by length across all passages into $SC(a)$ in a descending order.
\State $a_{\text{marked}}$: A dictionary that maps the the letters of the answer $a$ to either "marked" or "unmarked".
\State $\text{IsMarked}(sc, a, a_{\text{marked}})$: Returns true if the string $sc$ has already been marked or marked partially inside answer $a$, and false otherwise.
\State $\text{Mark}(sc, a, a_{\text{marked}})$: Marks the string in the answer as "marked" under passage $sc_p$.
\For{$sc \in SC(a)$}
    \If { $!\text{IsMarked}(sc, a, a_{\text{marked}})$} 
        \State $\text{Mark}(sc, a, a_{\text{marked}})$
    \EndIf
\EndFor
\State Format marked spans stored in $a_{\text{marked}}$ with the answer $a$ to form $a_{\text{span}}$.
\State \Return $a_{\text{span}}$
\end{algorithmic}
\end{algorithm}

\subsection{Method Comparison}

For each of the tasks described in \cref{sec:cite_task_explain}, we utilize every CoT method described in \cref{sec:cot_method_explain}. 
We run each combination with GPT-4, a small decoder-only model, and an encoder-decoder model.   
All results are presented on the test sets of the datasets mentioned in \cref{sec:dataset_names}.
For full implementation details of the training and inference, we refer the reader to section \ref{sec:impl_det} in the Appendix.

\setlength{\tabcolsep}{3pt}

\begin{table}[t]
\small
\centering
\begin{tabular}{llrrrrrr}
\toprule
\multirow[l]{2}{*} & \multirow[l]{2}{*}{CoT} &  \multicolumn{3}{c}{Answer Quality} & \multicolumn{3}{c}{Citation Quality} \\
\cmidrule(lr){3-5} \cmidrule(lr){6-8}
 &  & Span & Sent. & Pass. & Span & Sent. & Pass. \\
  \midrule
 & \multicolumn{7}{c}{QuoteSUM}  \\
     \cmidrule(lr){2-8}   
\multirow[c]{4}{*}{GPT-4} & None & 75.1 & 72.0 & 71.8 & 77.8 & 75.8 & \bfseries 68.3 \\
 & Span & \bfseries 75.9 & {72.9} & {72.8} & \bfseries 81.9 & 76.9 & 67.1 \\
 & Sent. & {75.8} & \bfseries 74.6 & \bfseries 73.2 & 77.8 & \bfseries 79.1 & 66.5 \\
 & Pass. & 75.5 & 72.4 & 72.0 & {80.6} & {77.3} & {67.7} \\

\midrule
\multirow[c]{4}{*}{Mistral 7B} & None & 74.2 & 72.6 & 74.4 & \bfseries 83.0 & 72.7 & \bfseries 66.4 \\
 & Span & \bfseries 76.2 & {75.6} & {75.1} & {82.2} & 72.3 & 64.7 \\
 & Sent. & 75.2 & \bfseries 75.6 & 75.1 & 79.9 & {73.4} & 65.0 \\
 & Pass. & {75.6} & 74.8 & \bfseries 75.8 & 82.0 & \bfseries 73.5 & {66.1} \\

\midrule
\multirow[c]{4}{*}{Flan-T5} & None & 76.6 & 73.8 & {74.0} & 84.7 & 71.4 & {64.3} \\
 & Span & 77.3 & 71.8 & 70.6 & 84.1 & 69.7 & 62.1 \\
 & Sent. & {78.4} & \bfseries 78.2 & \bfseries 76.7 & {84.7} & \bfseries 79.7 & \bfseries 66.5 \\
 & Pass. & \bfseries 78.5 & {74.4} & 73.1 & \bfseries 85.6 & {72.7} & 61.2 \\

  \midrule[0.9pt]
 & \multicolumn{7}{c}{MSMARCO}  \\
       \cmidrule(lr){2-8} 
\multirow[c]{4}{*}{Mistral 7B} & None & 69.6 & 69.1 & 70.2 & 68.4 & 66.1 & 66.0 \\
 & Span & {70.8} & 70.5 & {71.5} & \bfseries 74.8 & 67.8 & {68.9} \\
 & Sent. & \bfseries 71.0 & \bfseries 71.6 & 71.3 & 72.4 & {69.1} & 68.4 \\
 & Pass. & {70.8} & \bfseries 71.6 & \bfseries 71.6 & {74.4} & \bfseries 69.2 & \bfseries 69.1 \\

\midrule
\multirow[c]{4}{*}{Flan-T5} & None & {70.4} & 71.5 & {72.4} & {74.3} & 67.7 & 68.5 \\
 & Span & {70.4} & 70.3 & 68.4 & \bfseries 74.7 & 68.7 & 68.6 \\
 & Sent. & \bfseries 71.3 & {71.6} & 71.9 & 73.2 & \textbf{69.0} & {69.3} \\
 & Pass. & 69.0 & \bfseries 72.7 & \bfseries 72.6 & 73.9 & \bfseries 69.0 & \bfseries 70.0 \\

\end{tabular}
\caption{Average answer and citation quality results, per citation level, per CoT method, for QuoteSUM and MSMARCO. \textbf{Columns:} Citation levels. \textbf{Rows:} CoT method used and metrics averages per level.}
\label{tab:summary_overall_semqa_all}
\end{table}

\setlength{\tabcolsep}{3pt}

\begin{table*}[t]
\small
\centering
\begin{tabular}{llll}
Idx. & Expected Answer & No-CoT & Span-CoT \\ \hline
75        & 
\begin{tabular}[c]{@{}l@{}} 
(2) \textcolor{teal}{[ 2 general paralysis} \\
\textcolor{teal}{of the insane (GPI) ]}
\end{tabular} & 
\begin{tabular}[c]{@{}l@{}} 
 GPI stands for \textcolor{teal}{[ 2 general} \\
 \textcolor{teal}{paralysis of the} \\
 \textcolor{teal}{insane ]}, \textcolor{red}{[ 3 Georgia Budget }\\
 \textcolor{red}{and Policy Institute ]}, and \textcolor{red}{[ 5} \\
 \textcolor{red}{Global Residence Program Index ]}.
\end{tabular} & 
\begin{tabular}[c]{@{}l@{}} 
Lets analyze the relevant spans \\
of information from the input \\
passages. \\
\textbf{From passage [2]}, the relevant  \\
spans are the following: \\
\textbf{* general paralysis of the insane} \\
Thus, the final answer is: \\
GPI stands for \textcolor{teal}{[ 2 general paralysis} \\ 
\textcolor{teal}{of the insane ]}.
\end{tabular}                                \\ \hline
941        & 
\begin{tabular}[c]{@{}l@{}} 
Some sources state that \\
\textcolor{teal}{[ 1 "Born on the Fourth } \\
\textcolor{teal}{of July" ]} lasts \textcolor{teal}{[ 1 for} \\
\textcolor{teal}{two hours and } \\
\textcolor{teal}{twenty-five minutes. ]}

\end{tabular} & 
\begin{tabular}[c]{@{}l@{}} 
\textcolor{red}{[ 2 "Born on the Fourth of July"} \\
\textcolor{red}{was written in Santa Monica, California, } \\

\textcolor{red}{during the fall of 1974 } \\
\textcolor{red}{in exactly one month, three } \\
\textcolor{red}{weeks and two days. ]}
\end{tabular} & 
\begin{tabular}[c]{@{}l@{}} 
... \\
\textcolor{teal}{[ 1 Born on the Fourth of July ]}  \\
is the title of \textcolor{red}{[ 1 Stone's movie ]},  \\
which runs \textcolor{teal}{[ 1 two hours and} \\
\textcolor{teal}{twenty-five minutes ]}. \\
\textcolor{red}{[ 2 "Born on the Fourth of July"} \\
\textcolor{red}{was written in Santa Monica, } \\

\textcolor{red}{California, during the fall of 1974 } \\
\textcolor{red}{in exactly one month, three } \\
\textcolor{red}{weeks and two days. ]}

\end{tabular}                                \\ \hline
\end{tabular}
\caption{Examples of answer results when utilizing the span CoT method, as opposed to not using CoT at all. The examples are taken from the QuoteSUM test set, using the finetuned Mistral-7B models in both cases. }
\label{tab:cot_example_compare}
\end{table*}

\section{Experimental Results}



To examine how CoT affects the model, we inspect the GPT-4 case, and provide a detailed overview of the metrics in \cref{tab:gpt4_res_few}. In the answer metrics, the differences between the various CoT methods and the standard run (shown as \textit{"None"}) are not significant, aside from the sentence level, where the sentence CoT method produces better scores across all metrics. For the citation quality metrics, the span and sentence level produce better scores than their counterparts when using the appropriate CoT methods. 
\par In particular, the CSCA  metric in the span level case showcases a significant advantage, mainly since the model is guided to produce the spans relevant for the answer in advance, thus allowing the model to accurately reference them. 
\par In Table \ref{tab:cot_example_compare}, we showcase some examples of how the span CoT method allows the model to better quote spans from the original text. The first example shows that it allows the model to avoid quoting irrelevant spans. The second one showcases that while it does at times quote the wrong information, it still manages to quote the right information as well, as opposed to not using CoT at all.
\par These results also support our claims, as shown in the DOC F1 metric, showcasing a significant advantage for our method across the various citation levels. Since DOC F1 is lower for missing and excessive citations, our superior performance in it demonstrates how our approach improves upon those aspects.
At the passage level, the SEM-F1 and ALCE-F1 scores showcase an advantage for the standard run. We hypothesize that the addition of CoT for such a broad citation task is less helpful than in the more granular sentence and span level citations. 
We observe that the GPT-4 model performed best in the span level when using the span CoT method, and the same applies for the sentence level with sentence CoT. In the passage level, while the CoT methods improve the answer quality, they do not aid in the citation quality.

We showcase similar results for all the models, across the various citation levels and CoT methods in \cref{tab:finetune_quote,tab:finetune_msmarco}.
For each setting, we display all metrics and category averages for answer and citation quality, respectively.

We present an overview of the various combinations of citation levels and CoT methods in \cref{tab:summary_overall_semqa_all}. 
For smaller decoder-only models (i.e. Mistral 7B), the answer scores perform best in general when the citation level is used with its appropriate CoT method. In the citation metrics, results are different between the QSUM and MS. While the CoT methods behave well in the MS dataset, they are proving less helpful in QSUM, with the main difference being a higher ALCE-F1 score for the standard run. The difference can be mostly attributed to the higher complexity of the QSUM dataset, as opposed to the shorter and simpler MS answers. 

For encoder-decoder models, the CoT methods produce better results than the standard run, both in the answer and citation metrics. However, aside from the MS citation quality results, the advantage of using each citation level task with its corresponding CoT method is less prominent. The sentence CoT method is highly dominant in most of the metrics in the QSUM dataset, and passage CoT method being prominent in the MS dataset.

Finally, we observe in \cref{tab:summary_overall_semqa_all} that the smaller models are competitive with and even  outperform GPT-4. This conclusion is similar to the one reached by \citet{Schuster2023SEMQASM}, when comparing to other LLMs without finetuning.




\section{Related Work}
\label{sec:related_rerank}

Several recent studies have suggested different techniques for text generation including citation attribution, with varying degrees of detail. Attributing relevant information to web pages was done in \citet{thoppilan2022lamda,liuEvaluatingVerifiabilityGenerative2023, Bohnet2022AttributedQA}. \citet{menick2022teaching} attributes the response to specific snippets from the source passages. 
In particular, \citet{gao2023enabling} assigns attribution for each produced sentence to one or several input passages. Our sentence level citation approach and the evaluation metrics are based on this work.

Substantial work has been done to reduce hallucination in the generated text produced by the model \cite{li2024dawn,rawte2023exploring}, as citation has the ability to increase the reliability of the generated text.
Some work has been done on correlating answer to text sections in the retrieved passages \cite{gao2023enabling, bohnet2023attributed}. However, these techniques require additional processing.

The usage of CoT has shown significant improvements in a wide variety of tasks \cite{Wei2022ChainOT}.
Additionally, in-context learning can be added to the input, by providing examples of how to solve the task at hand \cite{wei2022emergent}. We explore the effects of using CoT with the various citation level tasks, in addition to the usage of in-context learning examples in every scenario. 
\citet{luo2023sail} instruct and finetune models to produce claims, based on retrieved passages, differentiating between informative and distracting input segments. We specify the text granularity level we expect the model should cite and phrase its answers.
\citet{slobodkin2024attribute} focus on the sentence citation task, splitting up the generation task into multiple reasoning steps, including CoT as well.
Our approach focuses on using a single call, producing the answer at-once, with a CoT approach.

\section{Conclusions}

Through extensive measurements applied across various models and citation levels, we have been able to ascertain the quality of both answers and citations. Our findings indicate that the use of our CoTAR reasoning significantly enhances the capacity of a model to generate superior quality answers and more accurate, faithful citations from the source. This improvement is evident in both decoder-only and encoder-decoder models. Furthermore, our research shows that by using finetuning, smaller models can compete with or even surpass the performance of GPT-4 across a wide range of answer and citation metrics. This is particularly evident when considering the diversity of citation levels and CoT methods.

\section{Limitations}
In our experimentation, we did not compare our approach directly to the work done in \citet{slobodkin2024attribute}, due to lack of access to the publicly available code-base.
In our GPT-4 runs, we ran the experimentation solely on the QuoteSUM dataset, in order to compare GPT-4 to our models on an previously established dataset. 
When considering datasets for our work, we primarily focused on the QuoteSUM dataset from \citet{Schuster2023SEMQASM}, and also the MSMARCO dataset, due to the extensive annotation of the answers and the accompanying passages. For future work, we will add similar datasets in structure and content.

\section{Potential Risks}

Since this project is primarily focused on assessing current models using datasets that are openly accessible, we do not foresee any possible adverse effects.

\bibliography{acl_anthology,references}
\appendix
\newpage
\section{Implementation Details}
\label{sec:impl_det}

The prompt is built using 4 fewshot examples, each composed of a question, a list of passages, and the expected answer in the task's format. As done by \citet{Schuster2023SEMQASM}, the prompt for the models is the same as they have specified, with an example in \cref{tab:prompt_span_level_example}, and the fewshot examples are chosen with the question similarity between the test question and the training example questions. As stated above, the expected answers cite their passages of origin in one of the levels described in \cref{tab:cite_level_example}. For the CoT setting, a prefix is attached in one of the methods showcased in \cref{tab:cot_level_prompts}.
Text generation is done without sampling, with a maximum length of 2k tokens. For QuoteSUM, the train and test sets are identical to the ones used by \citet{Schuster2023SEMQASM}. For MSMARCO, we utilize 2750 samples from the created dataset for training, and a 1000 for the test set. We specify the training parameters in \cref{tab:model_training_params}.

\section{Models}
\label{sec:models}

The weights for the models we have used have all been retrieved from \href{https://huggingface.co/models}{HuggingFace}, aside from GPT-4, for which we use Azure OpenAI:

\begin{itemize}[leftmargin=*]
\item Mistral 7B:  \href{https://huggingface.co/mistralai/Mistral-7B-Instruct-v0.2}{mistralai/Mistral-7B-Instruct-v0.2}, distributed under the \href{https://choosealicense.com/licenses/apache-2.0/}{Apache License 2.0}.
\item Flan-T5: \href{https://huggingface.co/google/flan-t5-xxl}{google/flan-t5-xxl}, distributed under the \href{https://choosealicense.com/licenses/apache-2.0/}{Apache License 2.0}. 
\item GPT-4: API access using Azure OpenAI, model \verb|gpt-4-32k-0613|.
\end{itemize}

\section{Additional Results}

\begin{table}[H]
\centering
\small
\begin{tabular}{@{}lc@{}c@{}}
\toprule
Parameter       & Decoder-Only & Encoder-Decoder             \\ \midrule
Max Target Len. & - & 4096               \\ 
Max Seq. Len. & 8192 & 32768               \\ 
Lora R & 16 & 16 \\
Lora $\alpha$ & 32 & 32 \\
Lora Dropout & 0.05 & 0.05 \\
Lora Bias & None & None \\
Lora Modules &  gate, down, up, $q$, $v$, $k$, $o$ & $q,v$ \\
LR              & 5e-5 & 2e-4              \\
LR Scheduler    & Linear & Linear            \\
Weight Decay    & 0 & 0.01              \\
Precision       & bfloat16 & bfloat16    \\
Batch Size      & 1 & 1                \\
Epochs  & 3 & 3             \\ 
Warmup Ratio    & 0 & 0.1              \\ \bottomrule
\end{tabular}
\caption{The training parameters used for Lora finetuning for the decoder-only and encoder-decoder models.}
\label{tab:model_training_params}
\end{table}

\begin{table*}[ht]
\centering
\begin{tabular}{ll}
\hline
Citation Level & Cited Answer\\ \hline
Passage        & \begin{tabular}[c]{@{}l@{}}" Johnny Panic and the Bible of Dreams " is a song by the British \\ band Tears for Fears. They also have international hit singles \\ such as: "Mothers Talk", "Shout", "Everybody Wants to Rule the World", \\ "Head over Heels", and "I Believe" [1][5].\end{tabular}                                \\ \hline
Sentence       & \begin{tabular}[c]{@{}l@{}}" Johnny Panic and the Bible of Dreams " is a song by the British \\ band Tears for Fears [1]. They also have international hit singles \\ such as: "Mothers Talk", "Shout", "Everybody Wants to Rule the World", \\ "Head over Heels", and "I Believe" [5].\end{tabular}                        \\ \hline
Span           & \begin{tabular}[c]{@{}l@{}}" {[} 1 Johnny Panic and the Bible of Dreams {]} " {[} 1 is a song by the British \\ band Tears for Fears {]} . They also have {[} 5 international hit singles {]}  \\ such as: {[} 5 "Mothers Talk", "Shout", "Everybody Wants to Rule the World", \\ "Head over Heels", and "I Believe". {]} \end{tabular} \\ \hline
\end{tabular}
\caption{Citation Level examples, with each row highlighting the approach for citing the source passages.}
\label{tab:cite_level_example}
\end{table*}

\begin{table*}[!ht]
\centering
\begin{tabular}{ll}
CoT Method & Answer Prefix\\ \hline
Passage        & \begin{tabular}[c]{@{}l@{}}Lets analyze the input passages. \\
The only relevant passages to the question are passages 1, 5. \\
Thus, the final answer is:
\end{tabular}                                \\ \hline
Sentence       & \begin{tabular}[c]{@{}l@{}}Lets analyze the relevant information from the input passages. \\
From passage [1], we know that: " Johnny Panic and the Bible of Dreams " \\ is a song by the British band Tears for Fears . \\
From passage [5], we know that: They also have international hit singles \\ such as: "Mothers Talk", "Shout", "Everybody Wants to Rule the World", "Head \\ over Heels", and "I Believe". \\
Thus, the final answer is:
\end{tabular}                        \\ \hline
Span           & \begin{tabular}[c]{@{}l@{}}Lets analyze the relevant spans of information from the input passages. \\
From passage [1], the relevant spans are the following: \\
  * Johnny Panic and the Bible of Dreams \\
  * is a song by the British band Tears for Fears \\
From passage [5], the relevant spans are the following: \\
  * international hit singles \\
  * "Mothers Talk", "Shout", "Everybody Wants to Rule the World", "Head \\ over Heels", and "I Believe". \\
Thus, the final answer is:
\end{tabular} \\ \hline
\end{tabular}
\caption{Examples of Chain-of-Thought prompt prefix for the various Levels. The model is instructed/finetuned to produce the snippets per CoT method.}
\label{tab:cot_level_prompts}
\end{table*}

\begin{table*}[ht]
\centering
\begin{tabular}{ll}
\hline
Answer the question by summarizing the given sources while explicitly copying \\ 
spans from the sources. When copying a span, use brackets and the respective \\ source number to indicate that this span was copied. Use explicit copying \\ as much as possible and for all factual statements, while preserving fluency. \\ Make sure to use all relevant sources and properly quote them. Here \\ 
are some examples: \\ 
Question: how much power does a wind turbine produce? \\ 
{[}1{]} Compact wind acceleration turbine: It is generally thought that since... \\
{[}2{]} Sustainable architecture: roof ledge. Small-scale rooftop wind turbines have ...\\
{[}3{]} Turby wind turbine: can because horizontal axis (HAWT) types cannot change ...\\
Quoted summary: One source states the {[} 1 amount of power produced by a wind \\ turbine is proportional to 
the cube of the wind speed {]} . Other sources state \\ {[} 2 Turbines for residential scale use {]} {[} 2 produce electricity at a rate of 900 watts to \\ 
10,000 watts {]} , and {[} 3 is specified to generate power in winds of \\ 
between 4 m/s (9 mph, 7.8kts) and 14 m/s (31 mph, 27.2kts) {]} . \\ 
\\
Question: a component is what? \\ 
{[}1{]} Modular programming: in Dart, Go or Java) is sometimes used instead of ...\\
{[}2{]} Physical body: the system at a point in time changes from identifying the object to... \\
Quoted summary: A {[} 1 component is a piece of a whole system {]} . Also, {[} 2 A \\ 
component is an object completely within the boundary of a containing object. {]} \\ 
\\
Question: what is the title of tears for fears song? \\ 
{[}1{]} Johnny Panic and the Bible of Dreams (song): ...\\
{[}2{]} Mark Crew: Will Ruin Your Life. The album was produced at his studio. ...\\
{[}3{]} Raoul and the Kings of Spain: Raoul and the Kings of Spain is the fifth studio ...\\
{[}4{]} Everybody Loves a Happy Ending: Everybody Loves a Happy Ending ...\\
{[}5{]} Songs from the Big Chair: Songs from the Big Chair is the second studio album ...\\
Quoted summary: Song titles by {[} 1 the British band Tears for Fears {]} include the \\ {[} 5 1985 {]} {[} 5 international hit singles "Mothers Talk", "Shout", "Everybody Wants to \\ Rule the World", "Head over Heels", and "I Believe" {]} , {[} 4 1989's "The Seeds of Love" {]} \\ , the {[} 1 1990 single "Advice for the Young at Heart" {]} with {[} 1 "Johnny Panic and the Bible \\ of Dreams" {]} on {[} 1 the B-side {]} , and " {[} 2 Lemon To A Knife Fight {]} ", \\ 
 " {[} 2 Cheetah Tongue {]} ", and " {[} 2 Turn {]} " from {[} 2 2018 {]} \\

\end{tabular}
\caption{Citation Level prompt example, containing 2 fewshot examples, with the relevant passages for each question provided.}
\label{tab:prompt_span_level_example}
\end{table*}

\begin{table*}[ht]
    \centering
\begin{tabular}{lllrrrrrrrrr}
\toprule

  \multirow[c]{2}{*}{Level} & \multirow[c]{2}{*}{CoT} & \multirow[c]{2}{*}{Model} &  \multicolumn{4}{c}{Answer Quality} & \multicolumn{5}{c}{Citation Quality} \\
  \cmidrule(lr){4-7} \cmidrule(lr){8-12}
 &  &  & BERT & HEM & RL & \textit{AVG.} & ALCE F1 & CSCA & DOC F1 & SEM-F1 & \textit{AVG.} \\
\midrule
\multirow[c]{12}{*}{Span} &  & GPT-4 & 91.71 & 67.58 & 66.08 & 75.12 & 77.06 & 81.15 & 85.33 & 67.57 & 77.78 \\
 & None & Flan & \bfseries 92.20 & 65.41 & 72.12 & 76.58 & 85.92 & 94.26 & 86.71 & 71.94 & 84.71 \\
 &  & Mist. & 91.61 & 60.93 & 70.03 & 74.19 & \bfseries 86.95 & 94.97 & 84.69 & 65.27 & 82.97 \\
  \cmidrule(lr){2-12}
 &  & GPT-4 & 92.06 & 66.95 & 68.55 & 75.85 & 78.83 & 89.88 & 86.54 & 72.23 & 81.87 \\
 & Span & Flan & 92.06 & 67.97 & 71.99 & 77.34 & 81.47 & 95.36 & 86.51 & 72.86 & 84.05 \\
 &  & Mist. & 92.04 & 65.98 & 70.67 & 76.23 & 74.60 & \bfseries 96.17 & 85.87 & 72.19 & 82.21 \\
    \cmidrule(lr){2-12}
 &  & GPT-4 & 91.87 & 66.88 & 68.76 & 75.84 & 84.76 & 69.69 & 86.38 & 70.52 & 77.84 \\
 & Sent. & Flan & 92.15 & \bfseries 71.46 & 71.64 & 78.42 & 86.71 & 90.02 & 87.79 & 74.41 & 84.73 \\
 &  & Mist. & 92.01 & 63.32 & 70.22 & 75.19 & 73.56 & 90.86 & 85.40 & 69.69 & 79.88 \\
    \cmidrule(lr){2-12}
 &  & GPT-4 & 91.91 & 66.89 & 67.67 & 75.49 & 79.21 & 86.94 & 86.49 & 69.87 & 80.63 \\
 & Pass. & Flan & 92.17 & 70.84 & \bfseries 72.47 & \bfseries 78.49 & 83.36 & 94.14 & \bfseries 88.01 & \bfseries 77.07 & \bfseries 85.64 \\
 &  & Mist. & 91.88 & 63.90 & 70.91 & 75.56 & 80.14 & 94.12 & 85.03 & 68.63 & 81.98 \\
\midrule[1.1pt]
\multirow[c]{12}{*}{Sent.} &  & GPT-4 & 91.47 & 62.48 & 62.18 & 72.04 & 82.79 &-& 85.32 & 59.20 & 75.77 \\
 & None & Flan & 91.84 & 60.43 & 69.22 & 73.83 & 79.37 &-& 81.18 & 53.66 & 71.40 \\
 &  & Mist. & 91.56 & 58.51 & 67.63 & 72.57 & 82.15 &-& 82.10 & 54.00 & 72.75 \\
    \cmidrule(lr){2-12}
 &  & GPT-4 & 91.92 & 62.06 & 64.76 & 72.91 & 82.39 &-& 86.71 & 61.74 & 76.94 \\
 & Span & Flan & 91.43 & 57.43 & 66.60 & 71.82 & 71.33 &-& 83.19 & 54.46 & 69.66 \\
 &  & Mist. & 91.93 & 65.04 & 69.69 & 75.55 & 73.14 &-& 84.48 & 59.29 & 72.30 \\
    \cmidrule(lr){2-12}
 &  & GPT-4 & 91.96 & 64.59 & 67.36 & 74.64 & \bfseries 86.42 &-& 86.51 & 64.44 & 79.12 \\
 & Sent. & Flan & \bfseries 92.37 & \bfseries 69.36 & \bfseries 72.87 & \bfseries 78.20 & 85.38 &-& \bfseries 87.03 & \bfseries 66.77 & \bfseries 79.72 \\
 &  & Mist. & 91.94 & 64.45 & 70.40 & 75.60 & 75.48 &-& 84.81 & 59.97 & 73.42 \\
    \cmidrule(lr){2-12}
 &  & GPT-4 & 91.69 & 61.19 & 64.42 & 72.43 & 84.25 &-& 85.92 & 61.79 & 77.32 \\
 & Pass. & Flan & 91.64 & 62.97 & 68.62 & 74.41 & 72.57 &-& 85.05 & 60.43 & 72.68 \\
 &  & Mist. & 91.99 & 62.90 & 69.37 & 74.76 & 75.39 &-& 85.18 & 60.07 & 73.55 \\
\midrule[1.1pt]
\multirow[c]{12}{*}{Pass.}  &  & GPT-4 & 91.55 & 61.13 & 62.76 & 71.81 & \bfseries 75.38 &-& 85.19 & \bfseries 44.31 & \bfseries 68.29 \\
 & None & Flan & 91.69 & 62.10 & 68.26 & 74.02 & 69.33 &-& 83.31 & 40.41 & 64.35 \\
 &  & Mist. & 91.74 & 63.64 & 67.91 & 74.43 & 75.30 &-& 83.93 & 40.08 & 66.44 \\
    \cmidrule(lr){2-12}
 &  & GPT-4 & 91.90 & 61.66 & 64.70 & 72.75 & 71.60 &-& 86.17 & 43.41 & 67.06 \\
 & Span & Flan & 91.24 & 54.87 & 65.56 & 70.56 & 65.51 &-& 81.98 & 38.76 & 62.09 \\
 &  & Mist. & 91.99 & 64.25 & 69.14 & 75.12 & 70.65 &-& 83.52 & 39.78 & 64.65 \\
    \cmidrule(lr){2-12}
 &  & GPT-4 & 91.84 & 62.43 & 65.32 & 73.20 & 70.63 &-& \bfseries 86.41 & 42.59 & 66.54 \\
 & Sent. & Flan & \bfseries 92.43 & \bfseries 65.17 & \bfseries 72.61 & \bfseries 76.73 & 73.00 &-& 84.47 & 42.05 & 66.51 \\
 &  & Mist. & 92.14 & 62.54 & 70.59 & 75.09 & 69.92 &-& 84.53 & 40.61 & 65.02 \\
    \cmidrule(lr){2-12}
 &  & GPT-4 & 91.69 & 60.40 & 63.88 & 71.99 & 73.28 &-& 85.64 & 44.12 & 67.68 \\
 & Pass. & Flan & 91.32 & 61.82 & 66.14 & 73.09 & 58.18 &-& 86.16 & 39.20 & 61.18 \\
 &  & Mist. & 92.15 & 64.54 & 70.83 & 75.84 & 72.92 &-& 84.64 & 40.76 & 66.11 \\
\bottomrule

\end{tabular}

\caption{Results for the GPT-4 with fewshot and the smaller fintuned models over the QuoteSUM dataset. Each row contains both the values of the metrics per citation level and CoT method, and the average score on per metric type on their right. Mist. denotes Misrtal-7B, and Flan denotes FlanT5-XXL.}
\label{tab:finetune_quote}
\end{table*}

\begin{table*}[ht]
    \centering
\begin{tabular}{lllrrrrrrrrr}
\toprule
  \multirow[c]{2}{*}{Level} & \multirow[c]{2}{*}{CoT} & \multirow[c]{2}{*}{Model} &  \multicolumn{4}{c}{Answer Quality} & \multicolumn{5}{c}{Citation Quality} \\
  \cmidrule(lr){4-7} \cmidrule(lr){8-12}
 &  &  & BERT & HEM & RL & \textit{AVG.} & ALCE F1 & CSCA & DOC F1 & SEM-F1 & \textit{AVG.} \\
\midrule
\multirow[c]{8}{*}{Span} & \multirow[c]{2}{*}{None} & Flan & 91.91 & 62.17 & 57.08 & 70.39 & \bfseries 96.97 & 91.88 & 64.70 & 43.80 & 74.34 \\
 &  & Mist. & 91.24 & 64.16 & 53.28 & 69.56 & 80.61 & 94.51 & 59.86 & 38.47 & 68.36 \\
\cmidrule(lr){2-12}
 & \multirow[c]{2}{*}{Span} & Flan & 91.80 & 62.83 & 56.50 & 70.38 & 95.69 & 93.58 & 64.90 & 44.72 & 74.72 \\
 & & Mist. & 91.88 & 63.77 & 56.63 & 70.76 & 93.41 & \bfseries 96.47 & 65.30 & 43.84 & \bfseries 74.75 \\
  \cmidrule(lr){2-12}
&  \multirow[c]{2}{*}{Sent.} & Flan & \bfseries 92.25 & 62.72 & \bfseries 58.80 & \bfseries 71.26 & 95.67 & 86.94 & \bfseries 65.50 & \bfseries 44.75 & 73.22 \\
&  & Mist. & 91.83 & \bfseries 64.31 & 56.85 & 70.99 & 93.38 & 91.62 & 63.40 & 41.18 & 72.40 \\
  \cmidrule(lr){2-12}
&  \multirow[c]{2}{*}{Pass.} & Flan & 91.90 & 56.24 & 58.77 & 68.97 & 95.93 & 90.09 & 64.90 & 44.55 & 73.87 \\
 &  & Mist. & 91.90 & 63.69 & 56.73 & 70.77 & 93.73 & 95.36 & \bfseries 65.50 & 43.05 & 74.41 \\
\midrule[1.1pt]
\multirow[c]{8}{*}{Sent.} & \multirow[c]{2}{*}{None} & Flan & 92.83 & 62.77 & 58.78 & 71.46 & 96.53 &-& 62.00 & 44.49 & 67.68 \\
 &  & Mist. & 92.06 & 60.26 & 54.90 & 69.08 & 91.67 &-& 63.62 & 42.99 & 66.09 \\
    \cmidrule(lr){2-12}
 & \multirow[c]{2}{*}{Span} & Flan & 92.37 & 62.20 & 56.42 & 70.33 & 95.92 &-& 65.30 & 44.74 & 68.65 \\
 &  & Mist. & 92.54 & 61.66 & 57.24 & 70.48 & 94.75 &-& 64.50 & 44.09 & 67.78 \\
    \cmidrule(lr){2-12}
 & \multirow[c]{2}{*}{Sent.} & Flan & 92.81 & 63.31 & 58.82 & 71.65 & 96.55 &-& 65.15 & 45.29 & 69.00 \\
&  & Mist. & 92.59 & 64.11 & 58.04 & 71.58 & \bfseries 96.95 &-& 65.45 & 45.01 & 69.14 \\
    \cmidrule(lr){2-12}
 & \multirow[c]{2}{*}{Pass.} & Flan & \bfseries 92.92 & \bfseries 65.69 & \bfseries 59.45 & \bfseries 72.69 & 96.73 &-& 64.70 & \bfseries 45.59 & 69.01 \\
&  & Mist. & 92.67 & 63.88 & 58.20 & 71.58 & 96.05 &-& \bfseries 66.40 & 45.14 & \bfseries 69.19 \\
\midrule[1.1pt]
\multirow[c]{8}{*}{Pass.} & \multirow[c]{2}{*}{None} & Flan & 92.96 & 64.38 & \bfseries 59.93 & 72.42 & 95.20 &-& 64.30 & 45.94 & 68.48 \\
 &  & Mist. & 92.29 & 62.06 & 56.32 & 70.22 & 92.54 &-& 62.80 & 42.75 & 66.03 \\
    \cmidrule(lr){2-12}
 & \multirow[c]{2}{*}{Span} & Flan & 92.17 & 55.69 & 57.29 & 68.38 & 96.10 &-& 65.00 & 44.59 & 68.56 \\
 &  & Mist. & 92.67 & 63.72 & 58.04 & 71.48 & 95.40 &-& 66.40 & 44.77 & 68.86 \\
    \cmidrule(lr){2-12}
& \multirow[c]{2}{*}{Sent.} & Flan & 92.68 & 64.91 & 58.02 & 71.87 & 96.70 &-& 65.80 & 45.44 & 69.31 \\
 &  & Mist. & 92.49 & 64.12 & 57.16 & 71.26 & 96.30 &-& 64.80 & 44.20 & 68.43 \\
    \cmidrule(lr){2-12}
& \multirow[c]{2}{*}{Pass.} & Flan & \bfseries 92.97 & \bfseries 65.20 & 59.49 & \bfseries 72.55 & 96.60 &-& \bfseries 67.40 & \bfseries 46.10 & \bfseries 70.03 \\
 &  & Mist. & 92.75 & 64.03 & 58.14 & 71.64 & \bfseries 96.80 &-& 65.60 & 44.78 & 69.06 \\
\bottomrule
\end{tabular}

\caption{Results for the smaller finetuned models over the MSMARCO dataset. Each row contains both the values of the metrics per citation level and CoT method, and the average score on per metric type on their right. Mist. denotes Misrtal-7B, and Flan denotes FlanT5-XXL.}
\label{tab:finetune_msmarco}
\end{table*}

\end{document}